\documentclass[10pt,twocolumn,letterpaper]{article}

\usepackage{iccv}
\usepackage{times}
\usepackage{epsfig}
\usepackage{graphicx}
\usepackage{amsmath}
\usepackage{amssymb}

\usepackage[pagebackref=true,breaklinks=true,letterpaper=true,colorlinks,bookmarks=false]{hyperref}

\iccvfinalcopy 

\usepackage{subfigure}
\usepackage{multirow}
\usepackage[english]{babel}

\usepackage{booktabs,tabularx}
\usepackage{siunitx}
\newcolumntype{C}{>{\centering\arraybackslash}X}

\def\iccvPaperID{****} 

\ificcvfinal\fi

\begin{document}

\title{Condensing a Sequence to One Informative Frame for Video Recognition}

\author{Zhaofan Qiu$^{\dagger}$, Ting Yao$^{\dagger}$, Yan Shu$^{\S}$ Chong-Wah Ngo$^{\ddagger}$, and Tao Mei$^{\dagger}$\\
\parbox{20em}{\small\centering $^{\dagger}$ JD AI Research, Beijing, China}\\
\parbox{40em}{\small\centering $^{\S}$ University of Science and Techology of China, Hefei, China~~~~~~~~~~~~~~~~~~~$^{\ddagger}$ Singapore Management University, Singapore}\\
{\tt\small \{zhaofanqiu, tingyao.ustc, shuy.ustc\}@gmail.com, cwngo@smu.edu.sg, tmei@jd.com}
}

\maketitle
\ificcvfinal\thispagestyle{empty}\fi

\begin{abstract}
Video is complex due to large variations in motion and rich content in fine-grained visual details. Abstracting useful information from such information-intensive media requires exhaustive computing resources. This paper studies a two-step alternative that first condenses the video sequence to an informative ``frame'' and then exploits off-the-shelf image recognition system on the synthetic frame. A valid question is how to define ``useful information'' and then distill it from a video sequence down to one synthetic frame. This paper presents a novel Informative Frame Synthesis (IFS) architecture that incorporates three objective tasks, i.e., appearance reconstruction, video categorization, motion estimation, and two regularizers, i.e., adversarial learning, color consistency. Each task equips the synthetic frame with one ability, while each regularizer enhances its visual quality. With these, by jointly learning the frame synthesis in an end-to-end manner, the generated frame is expected to encapsulate the required spatio-temporal information useful for video analysis. Extensive experiments are conducted on the large-scale Kinetics dataset. When comparing to baseline methods that map video sequence to a single image, IFS shows superior performance. More remarkably, IFS consistently demonstrates evident improvements on image-based 2D networks and clip-based 3D networks, and achieves comparable performance with the state-of-the-art methods with less computational cost.
\end{abstract}

\section{Introduction}
Recently, the development of Convolutional Neural Networks (CNN) convincingly demonstrates high capability of CNN in image-domain visual recognition. For instance, an ensemble of residual nets \cite{he2015deep} achieves 3.5\% top-5 error on the ImageNet test set, which is even lower than 5.1\% of the reported human-level performance. Nevertheless, it is not trivial to apply a 2D CNN for video recognition. Since video is a temporal sequence with large variations and complexities, performing 2D CNN on individual frame cannot model temporal evolution across frames.

\begin{figure}[!tb]
   \centering
   \subfigure[Late fusion]{
     \label{fig:intro:a}
     \includegraphics[width=0.24\textwidth]{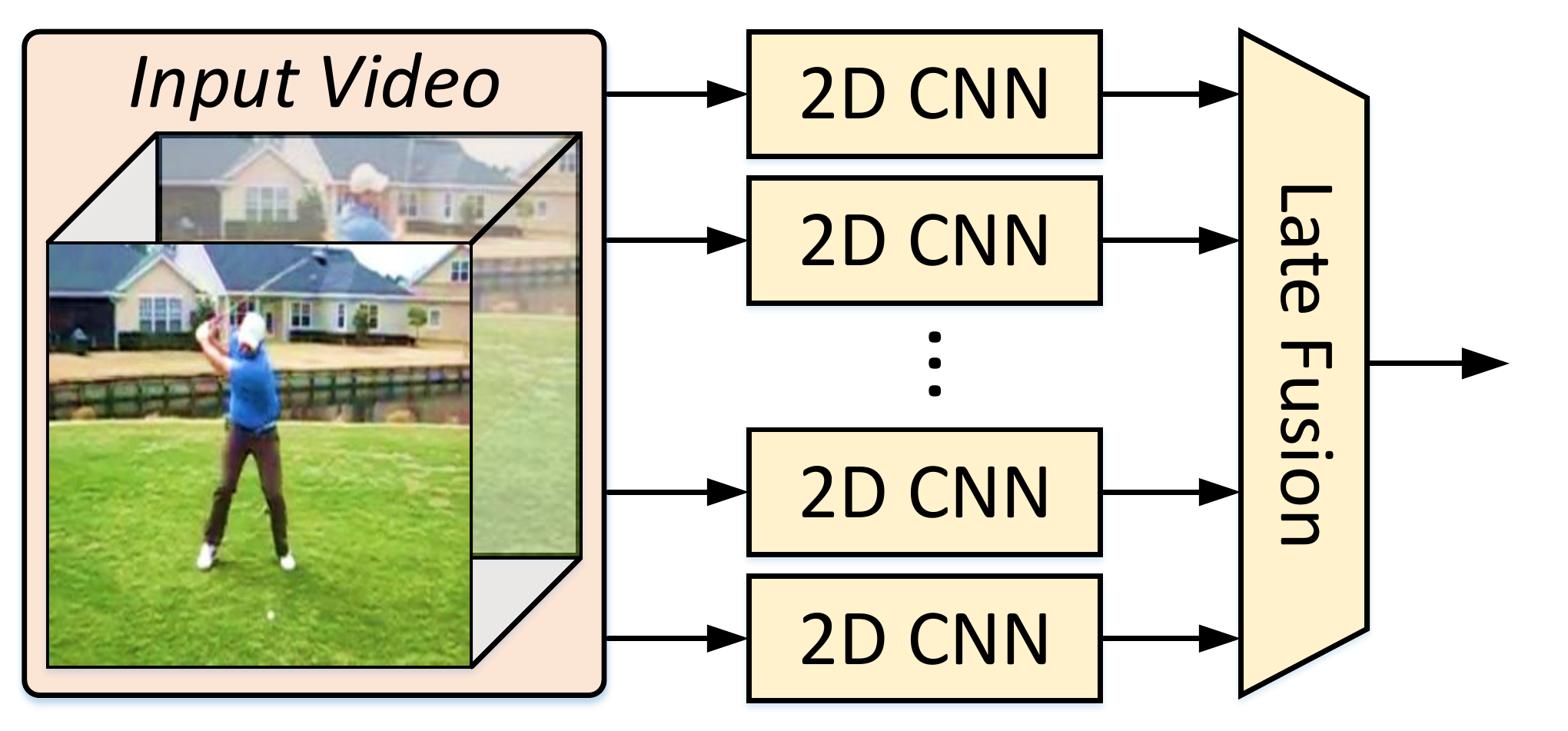}}
   \subfigure[Slow fusion]{
     \label{fig:intro:b}
     \includegraphics[width=0.212\textwidth]{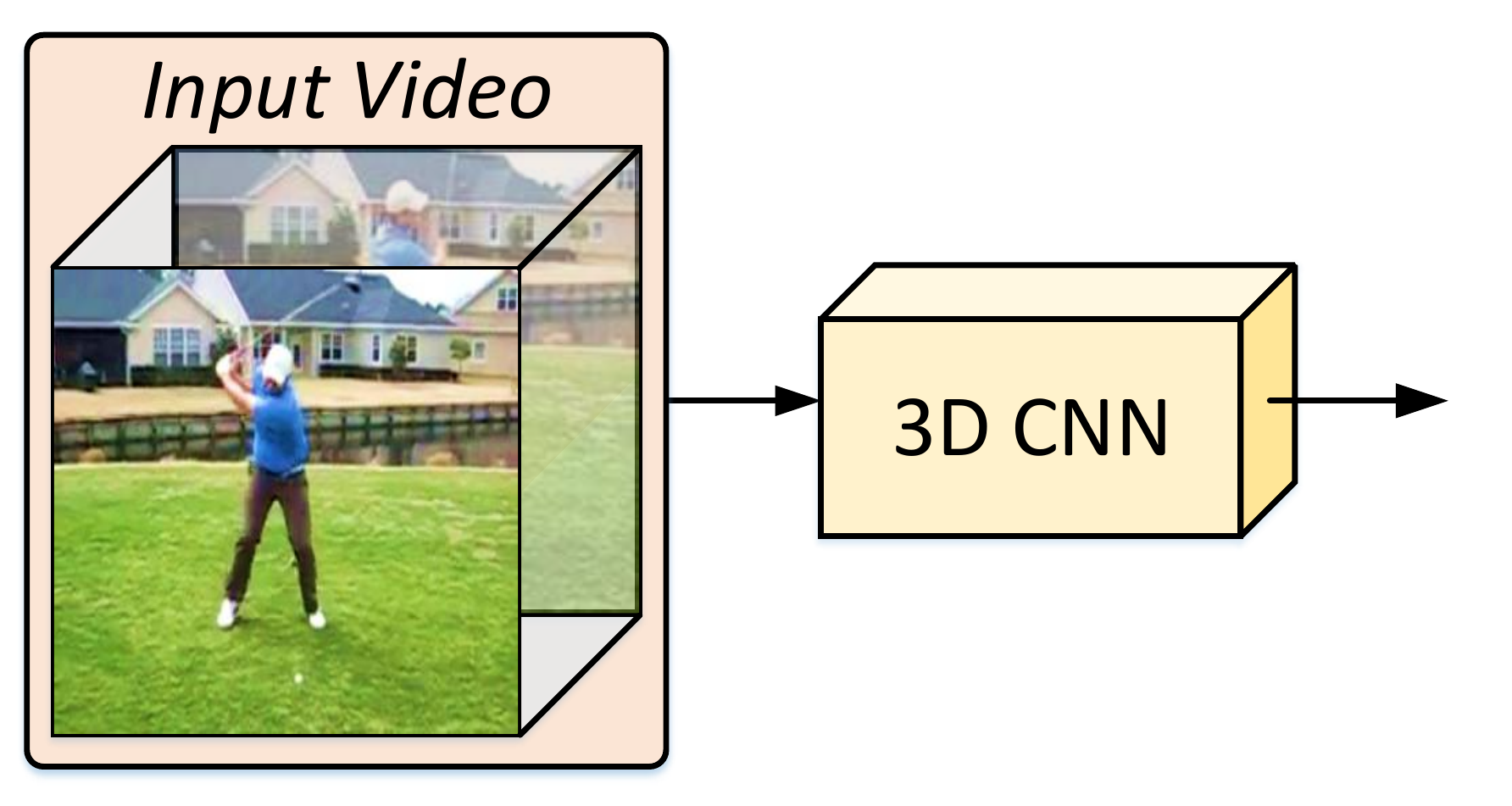}}
   \subfigure[Early fusion by frame synthesis]{
     \label{fig:intro:c}
     \includegraphics[width=0.435\textwidth]{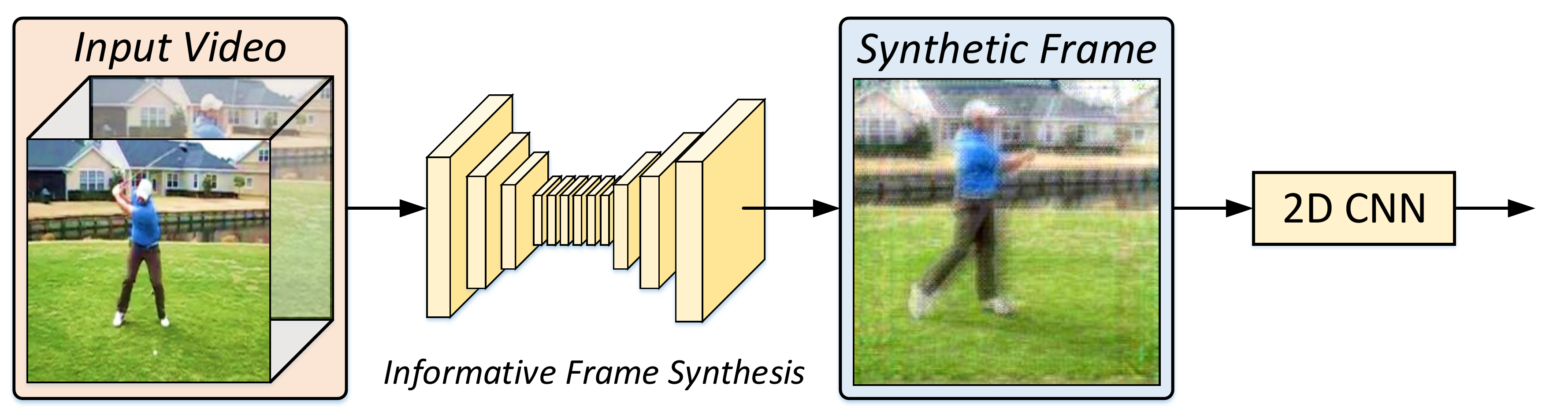}}
   \caption{\small Modeling temporal evolution for video recognition by (a) aggregating 2D representations from sampled frames by \textbf{late fusion}, (b) \textbf{slow fusing} the input frames by 3D CNN and (c) condensing the input video to one frame as \textbf{early fusion}.}
   \label{fig:intro}
   \vspace{-0.25in}
\end{figure}

Extensive progresses have been made to model the temporal sequence for video recognition. These works can be grouped into three categories according to which stage temporal information is aggregated, as shown in Figure \ref{fig:intro}. The first one is \textbf{late fusion} \cite{karpathy2014large,simonyan2014two,srivastava2015unsupervised,wang2018temporal,yue2015beyond}, which first extracts the image representation from 2D CNN for each frame and then aggregates the feature sequence for video recognition. Despite being straightforward by employing 2D CNN on video data, pixel-level temporal evolution over frames is overlooked. The second is \textbf{slow fusion} \cite{carreira2017quo,feichtenhofer2019slowfast,ji20133d,karpathy2014large,qiu2017learning,qiu2019learning,tran2015learning,tran2018closer}, which feeds entire video clip into a network (e.g., 3D CNN) for spatio-temporal convolution. This type of networks builds temporal connections among pixels across space and time at the expense of computational cost. For example, the de facto ResNet-101 \cite{he2015deep} requires 10G floating-number operations (FLOPs) for single crop on image data. When transferring this backbone to 3D CNN for 128-frame clip, the number of FLOPs is increased to 234G for SlowFast networks \cite{feichtenhofer2019slowfast}.

This paper addresses a new direction that \textbf{early fuses} the information from video sequence to one synthetic frame. Despite being a 2D image, the frame captures motion dynamics and visual details of a sequence. In this way, 2D CNN can be employed to learn both visual appearance as well as temporal evolution from just one frame. To this end, the crucial issue of early fusion becomes what information deserves to be preserved in a synthetic frame. Therefore, we propose a novel Informative Frame Synthesis (IFS) architecture guiding the generation of synthetic frames by multiple pre-defined tasks and regularizers. This architecture mainly consists of three components, i.e., a convolutional encoder-decoder network to transfer the input video sequence to a synthetic frame, three objective tasks to ensure that the frame is reconstructable, semantically and dynamically consistent with the original sequence, and two regularizers to preserve the fine-grained visual details. By jointly optimizing the transformation network in an end-to-end manner, the generated frame will attempt to encapsulate the required information of each task. In this way, the frame captures temporal evolution for video applications.

The main contribution of this work can be summarized as followings. First, IFS network is novelly proposed to learn the transformation from 3D video clips to 2D image frame. Second, different objectives and regularizers are proposed to encapsulate motion dynamics and visual details in a 2D frame. Extensive experiments are conducted on Kinetics dataset. Ablation studies investigate the impact of each task and regularizer towards frame synthesis. The results demonstrate that IFS with 2D CNN (e.g., ResNet-101) achieves comparable performance to more computationally expensive 3D CNN (e.g., I3D \cite{carreira2017quo}). When the synthetic frames are stacked as a video summary for 3D CNN classification, higher performance is attained than most of the existing works with less computation.

\section{Related Work}
Video recognition has attracted intensive research interests in recent years due to its importance in different application areas, such as video surveillance, indexing, retrieval and robotics. We briefly group the methods for video recognition into two categories: hand-crafted feature-based and deep learning-based methods.

Early progresses are mostly based on the classifiers trained on \textbf{hand-crafted feature}, which usually starts by detecting spatio-temporal interest points and then describing them with local representations. Examples of hand-crafted feature include Space-Time Interest Points (STIP) \cite{laptev2005space}, Histogram of Gradient and Histogram of Optical Flow \cite{laptev2008learning}, 3D HOG \cite{klaser2008spatio}, SIFT-3D \cite{scovanner20073}, Extended SURF \cite{willems2008efficient}, and improved dense trajectory \cite{wang2011action,wang2013action}. These hand-crafted descriptors are not particularly optimized for the video recognition task and may lack discriminative capacity.

The most recent approaches for video recognition are to devise \textbf{deep architectures} for end-to-end representation learning. Karparthy \emph{et al.} stack CNN-based frame-level representations in a fixed size of windows and then leverage spatio-temporal convolutions for video categorization \cite{karpathy2014large}. Benefiting from the usage of optical flow, in \cite{simonyan2014two}, the famous two-stream architecture is devised by applying two 2D CNN architectures separately on visual frames and stacked optical flows. Following the solution of two-stream networks, various schemes have been developed including convolutional fusion \cite{feichtenhofer2016convolutional}, key-volume mining \cite{zhu2016key}, temporal segment networks \cite{wang2018temporal} and temporal linear encoding \cite{diba2017deep}. To overcome the limitation of performing 2D CNN on modeling long-term dependencies, LSTM-RNN is proposed by Ng \emph et al. \cite{yue2015beyond} to model long-range temporal dynamics in videos. The aforementioned approaches are limited by treating video as a sequence of frames and optical flow for video understanding. More concretely, pixel-level temporal evolution across consecutive frames are not explored. The problem is addressed by 3D CNN proposed by Ji \emph{et al.} \cite{ji20133d}, which directly learns spatio-temporal representation from a short video clip. Later in \cite{tran2015learning}, Tran \emph{et al.} devise a widely adopted 3D CNN, namely C3D, for supervised learning of video representation over 16-frame video clips using large-scale video datasets. Furthermore, performance of the 3D CNN is further boosted by inflated 2D kernels \cite{carreira2017quo}, decomposed 3D kernels \cite{qiu2017learning,tran2018closer}, SlowFast networks \cite{feichtenhofer2019slowfast,wu2020multigrid} and depth-wise 3D convolutions \cite{duan2020omni,feichtenhofer2020x3d,tran2019video}.

Our work also falls into the category of deep architecture learning, but in a direction that is seldom explored. The proposed network aims to condense video sequence into a synthetic frame summarizing spatio-temporal evolution. An early work DI \cite{bilen2016dynamic} generates single ``dynamic image'' for each video by rank pooling technique to capture the temporal evolution. This pooling mechanism is improved in SVMP \cite{wang2018video} by multiple instance learning context and decision boundaries in SVM. More recently, AWSD \cite{tavakolian2019awsd} distills the video sequence to single image by weighted pooling the surrounding pixels in a local window with adaptive duration. These three works transform the video clip to single image by manually designed formulation and are not learnable. The most closely related work is AVD \cite{tavakolian2019avd}, which generates single image from video sequence by 3D CNN with adversarial learning.
Different from \cite{tavakolian2019avd}, our work focuses on the design of objective tasks and regularizers that can capture discriminative spatio-temporal characteristics in a single frame for recognition.

\begin{figure*}[!tb]
   \centering {\includegraphics[width=0.98\textwidth]{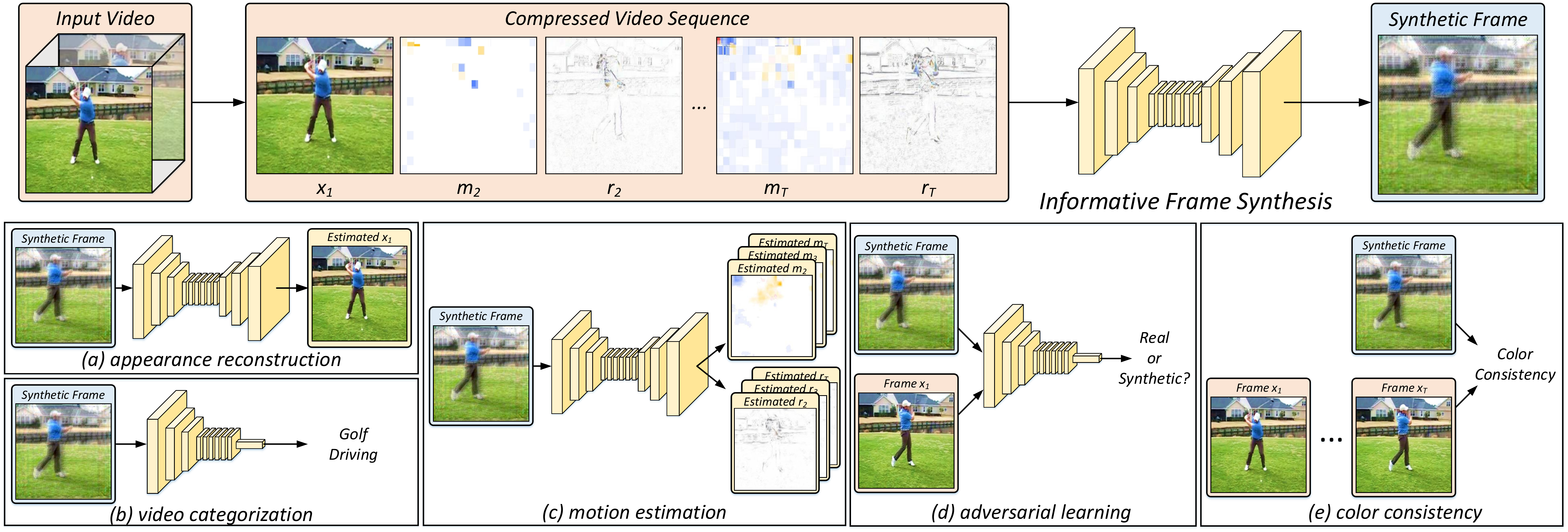}}
   \caption{\small The three main components of Information Frame Synthesis (IFS): the convolution encoder-decoder network for transferring of video clip to a synthetic frame (upper); three objectives driving 3D to 2D transformation (lower left); two regularizers improving the visual quality of synthetic frame (lower right).}
   \label{fig:framework}
   \vspace{-0.2in}
\end{figure*}

\section{Informative Frame Synthesis (IFS)}
IFS is essentially a generative model to synthesize a single frame that can infer visual and motion dynamics of a video clip. Figure \ref{fig:framework} depicts the overview of IFS. We begin this section by presenting the problem formulation, followed by definition of tasks and regularizers.
An end-to-end learning combining the loss functions is subsequently presented to train IFS. Finally, we explore different ways of utilizing IFS, including generating synthetic frames as a summary of long videos, for video recognition.

\subsection{Problem Formulation}
Denote a video clip with $T$ frames as $\mathcal{X}=\{x_{t}|t=1,...,T\}$, where each frame is in resolution of $H\times W$ and has $C$ color channels. Let IFS as a convolutional encoder-decoder network of function $\mathcal{F}$. Then, the problem of 2D frame synthesis is $\hat{x}=\mathcal{F}(\mathcal{X})$ where $\hat{x}\in \mathbb{R}^{C\times H \times W}$ is the 2D synthetic frame.
As successive frames are expected to be visually similar, we follow the typical way of video compression by presenting a frame as its difference w.r.t to a key frame.
Formally, take the first frame $x_{1}$ as the key frame, each subsequent frame $x_{t}$ can be viewed as a distorted key frame of $x_1$ by motion vector $m_{t}$ plus a residual $r_{t}$:
\begin{equation}\label{eq:compress}
\begin{aligned}
x_{t}=x_{1}\circ m_{t} + r_{t}
\end{aligned}~~,
\end{equation}
where $\circ$ denotes the pixel-level movement along the motion vector. In video compression, $x_1$ is named I-frame (intra-coded frame), and $\{m_{t}, r_{t}\}$ is the difference between the I-frame and a P-frame (predictive frame). Thus the video sequence $\mathcal{X}$ is reformulated as a compressed video sequence $\{x_1,m_2,r_2,...,m_T,r_T\}$ as in \cite{wu2018compressed,zhang2016real}, which can be directly extracted from the videos encoded with standards such as MPEG-4, H.264, HEVC, etc.

\subsection{Tasks for Synthetic Frame}
A synthetic frame $\hat{x}$ is expected to capture the most essential information required for video recognition. This includes the abilities of (1) reconstructing the appearance of input clip $\mathcal{X}$ from $\hat{x}$, (2) predicting the semantic category of $\mathcal{X}$, if available during training time, based on $\hat{x}$, and (3) estimating the temporal dynamics over frames. To this end, we optimize the encoder-decoder network $\mathcal{F}$ by jointly learning the following three~tasks.

\textbf{Appearance Reconstruction.} The synthetic frame $\hat{x}$ only contains $1/T$ digits compared with input clip. One straightforward way for video data distillation is to make the network $\mathcal{F}$ reversible. In other words, the input clip $\mathcal{X}$ can be recovered from the synthetic frame $\hat{x}$. In this task, we mainly focus on reconstructing the appearance of key frame. Considering the frame representation in compressed format as in Equ. (\ref{eq:compress}), only the first frame ($x_{1}$) of $\mathcal{X}$ is retained while other frames contain only the motion coefficients and residuals to predict $x_t$. Therefore, we design another convolutional encoder-decoder network $\mathcal{F}_{a}^{-1}$ aiming at recovering the first frame $x_{1}$ from $\hat{x}$. The objective function is given by the mean squared error (MSE) between $x_{1}$ and the recovered frame as
\begin{equation}\label{eq:loss-app}
\begin{aligned}
\mathcal{L}_{app}(\mathcal{F}, \mathcal{F}_{a}^{-1})=\left\| x_{1} - \mathcal{F}_{a}^{-1}(\mathcal{F}(\mathcal{X})) \right \|_{2}^{2}
\end{aligned}~~,
\end{equation}
where $\left\| \cdot \right\|_{2}^{2}$ denotes the mean squared $L2$ norm across the entire frame. 
The reconstruction loss enforces $\hat{x}$ to recall the initial appearance of $\mathcal{X}$.

\textbf{Video Categorization.}
Video label characterizes the spatio-temporal content of a video.
Here, we employ a convolutional encoder network $\mathcal{C}$ trying to predict the video category from synthetic frame. Given the supervised pair of video clip and label $\{\mathcal{X}, y\}$, the cross entropy loss measures the deviation between the ground truth and the predicted label from network $\mathcal{C}$ as
\begin{equation}\label{eq:loss2}
\begin{aligned}
\mathcal{L}_{cat}(\mathcal{F}, \mathcal{C})=\text{CrossEntropy}\{y, \mathcal{C}(\mathcal{F}(\mathcal{X}))\}
\end{aligned}~~,
\end{equation}
where $\mathcal{C}(\mathcal{F}(\mathcal{X}))$ is the probability of each category after being normalized to $(0, 1)$ by softmax operation. It is worth noticing that, among all the tasks and regularizers, video categorization is the only task that needs manual labels for supervised learning.

\textbf{Motion Estimation.}
Optical flow is usually extracted to describe the displacement between two consecutive frames, representing a transformation function that warps one frame to another. Similar in spirit, we try to empower the synthetic frame with the capacity to transfer the first input frame $x_1$ to the $t$-th frame $x_t$. Benefited from the compressed representation in Equ. (\ref{eq:compress}), this capacity is equivalent to reconstruct the motion vector $m_t$ and the residual $r_t$. These maps capture only the changes of video based on the information in I-frame. We design an identical convolutional encoder-decoder network $\mathcal{F}_{m}^{-1}$ to predict all the motion vectors and residuals from the synthetic frame $\hat{x}$ in one feed-forward propagation. Then, the loss for motion estimation task is calculated by the averaged MSE between the estimated motion vector/residual and the real motion vector/residual as
\begin{equation}\label{eq:loss-mot}
\begin{aligned}
&\{\hat{m}_2,\hat{r}_2,...,\hat{m}_T,\hat{r}_T\}=\mathcal{F}_{m}^{-1}(\mathcal{F}(\mathcal{X})), \\
&\mathcal{L}_{mot}(\mathcal{F}, \mathcal{F}_{m}^{-1})=\frac{1}{T-1}\sum_{t=2}^{T} \left\| m_{t} - \hat{m}_{t} \right \|_{2}^{2} + \left\| r_{t} - \hat{r}_{t} \right \|_{2}^{2}.
\end{aligned}
\end{equation}
Please note that there are $T-1$ motion vector maps and $T-1$ residual maps in total, altogether forming more digits than the synthetic frame. Such information ``bottleneck'' architecture will attempt to summarize the temporal dynamic across the entire sequence in a synthetic frame $\hat{x}$ by reducing inter-frame redundancy.

\subsection{Regularizers for Synthetic Frame}
The aforementioned three tasks drive the learning of the synthetic frame to recapitulate the appearance, semantic and motion information of input clip.
From a different perspective, two regularizers are designed to enhance the visual quality of the generated frame.

\textbf{Adversarial Learning.}
The first regularizer is derived from adversarial learning to force the synthetic frame to be visually similar to a real frame. A convolutional encoder network $\mathcal{D}$ is exploited as a discriminator to differentiate between the real and synthetic frames, while the network $\mathcal{F}$ is trained to maximally fool the discriminator by generating high-quality synthetic frames. The design follows generative adversarial learning \cite{Goodfellow:NIPS14} that trains two models, i.e., a generative model and a discriminative model, by pitting them against each other. The adversarial principle provides guidance for the frame synthesis by making the texture, pattern and structure between the real and synthetic frames indistinguishable by the discriminator. Formally, given the generated synthetic frame $\mathcal{F}(\mathcal{X})$ and the I-frame $x_1$, we calculate the adversarial loss as
\begin{equation}\label{eq:loss-adv}
\begin{aligned}
&\mathcal{R}_{adv}(\mathcal{D})=\left\| \mathcal{D}(x_1) \right\|_{2}^{2} + \left\|1-\mathcal{D}(\mathcal{F}(\mathcal{X}))\right\|_{2}^{2},\\
&\mathcal{R}_{adv}(\mathcal{F})=\left\|\mathcal{D}(\mathcal{F}(\mathcal{X}))\right\|_{2}^{2},
\end{aligned}
\end{equation}
where $\mathcal{D}(\cdot)$ denotes the score to measure the reality of a frame by discriminator network $\mathcal{D}$. The loss function used in Equ. (\ref{eq:loss-adv}) is the least-square GAN in \cite{zhu2017unpaired}, which performs more stably in joint training. Similar to the standard GANs, the training of adversarial learning in IFS is a minmax game between $\mathcal{F}$ and $\mathcal{D}$, expecting an good equilibrium that $\mathcal{F}$ can produce a ``realistic'' synthetic frame after convergence.

\textbf{Color Consistency.}
The second regularizer considers the meaning of each channel in a synthetic frame. Take the input video with RGB color space as an example, the number of channels $C$ is equal to 3 for the input and synthetic frame. However, the meaning of each channel and the correlation between channels in the synthetic frame are usually not constrained. Not surprisingly, without this constraint, we observe that the network $\mathcal{F}$ will easily converge to generate an unpredictable and stochastic color space. The hue information of video content will be lost in that color space. Therefore, we propose an efficient way to enhance the color information in the synthetic frame by minimizing the distance between the average RGB value in the input video and that in the synthetic frame. Specifically, the color consistency loss is defined as
\begin{equation}\label{eq:loss4}
\begin{aligned}
\mathcal{R}_{color}(\mathcal{F})=\frac{1}{T} \sum_{x_t \in \mathcal{X}}\left\| {\text{Ave}(x_t)}- \text{Ave}(\mathcal{F}(\mathcal{X})) \right\|_{2}^{2}
\end{aligned}~~,
\end{equation}
where $\text{Ave}(\cdot) \in \mathbb{R}^{C}$ denotes the mean value of each channel averaged across $H\times W$ positions inside the frame.

\subsection{Optimization} \label{sec:opt}
The overall training objective function of IFS integrates the losses from three tasks and two regularizers. The synthetic frame generation network $\mathcal{F}$ is updated as
\begin{equation}\label{eq:loss_all}
\begin{aligned}
\mathcal{L}=\mathcal{L}_{app} + \mathcal{L}_{cat} + \mathcal{L}_{mot} + \mathcal{R}_{adv} + \mathcal{R}_{color}
\end{aligned}~~,
\end{equation}
where the five losses are accumulated equally without weighting. Simultaneously, $\mathcal{F}_{a}^{-1}$, $\mathcal{C}$ and $\mathcal{F}_{m}^{-1}$ for the three tasks and $D$ for the adversarial regularizer are jointly optimized with $\mathcal{F}$ for their respective objectives.

\subsection{Video Recognition with Synthetic Frame} \label{sec:vr}
\begin{figure}[!tb]
   \centering
   \subfigure[Synthetic frame + 2D CNN]{
     \label{fig:injection:a}
     \includegraphics[width=0.47\textwidth]{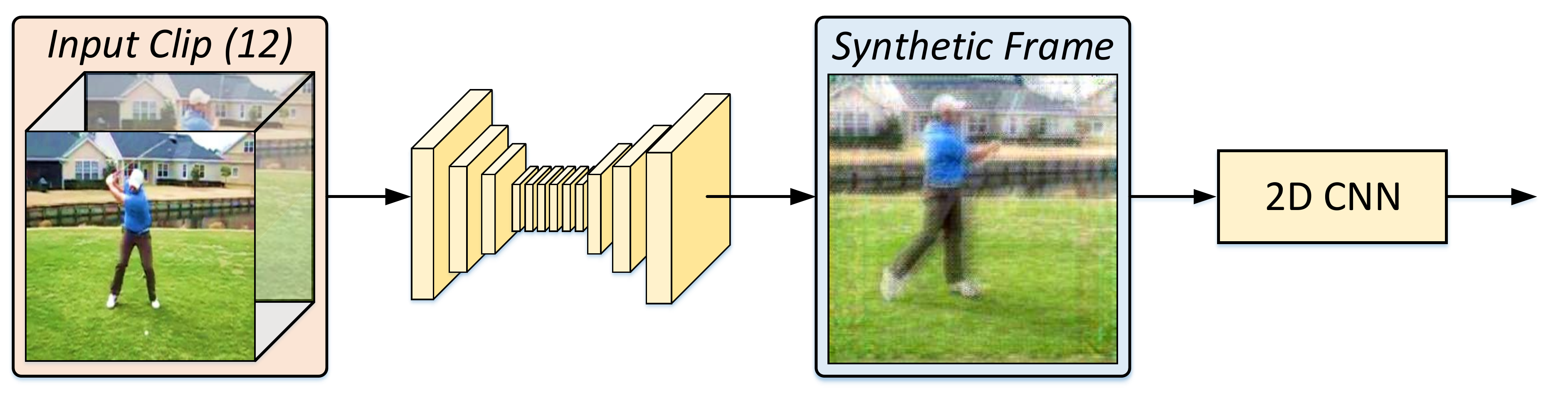}}
   \subfigure[Synthetic clip + 3D CNN]{
     \label{fig:injection:b}
     \includegraphics[width=0.47\textwidth]{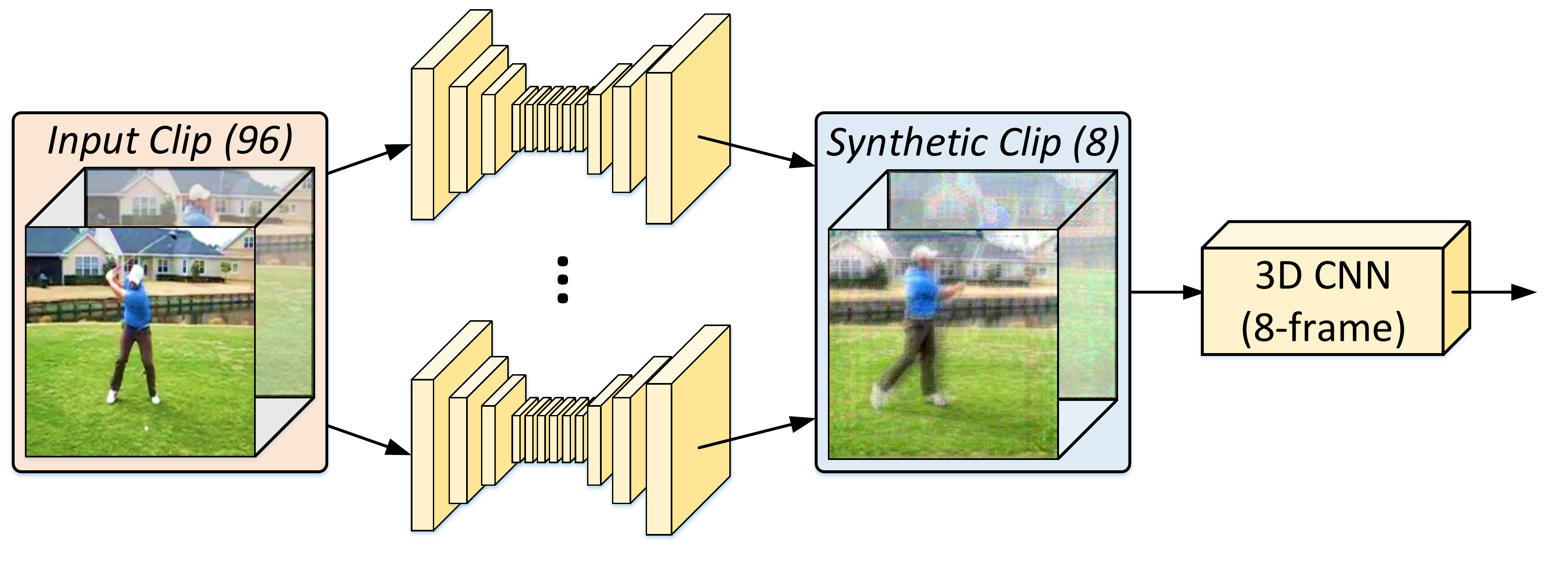}}
   \caption{\small Two examples of video recognition framework by integrating frame synthesis as early fusion.}
   \label{fig:injection}
   \vspace{-0.22in}
\end{figure}

We develop several video recognition frameworks by employing 2D CNN and 3D CNN respectively on the synthetic frame. Figure \ref{fig:injection} illustrates the two video recognition frameworks based on two typical networks. 

\vspace{0.07in}\noindent (i) \emph{Synthetic frame + 2D CNN:} The first one is simply by building a 2D CNN for the classification of each synthetic frame. Take $T=12$ (i.e., one I-frame plus 11 P-frames) as an example, the 2D CNN on synthetic frame plays a similar role as a 12-frame 3D CNN. We employ ResNet-101 \cite{he2015deep} pre-trained on ImageNet dataset \cite{ILSVRC15} as the 2D classifier. We refer the 2D CNN as the default video classifier in our experiments unless otherwise stated.

\vspace{0.07in}\noindent (ii) \emph{Synthetic clip + 3D CNN:} IFS can summarize a lengthy video into a short clip for video classification. Taking a 96-frame video as an example, IFS generates 8 consecutive synthetic frames, where each frame summarizes a 12-frame clip with a non-overlapping sliding window. In this case, we extend the 2D ResNet-101 to a 8-frame 3D CNN by the strategy in \cite{feichtenhofer2019slowfast,qiu2017learning,tran2018closer} that inserts one $3\times 1\times 1$ temporal convolution after each spatial convolution. Similar as the SlowFast networks \cite{feichtenhofer2019slowfast}, we do not perform temporal down-sampling in our 3D CNN, which means that the temporal dimension of each feature map is fixed as 8.

As shown in Figure \ref{fig:injection}, by early fusion of frames, IFS either condenses a video clip into one frame for 2D CNN classification or summarizes a clip as a synopsis of 8 frames for 3D CNN classification. The reduction is as much as 12 times of the original video length, which greatly decreases the computational cost.

IFS can also be degenerated to focus on capturing motion dynamics only by excluding the task of appearance construction. We name this variant as \textbf{IFS-mot}, where the produced synthetic frame only summarizes the information not included in $x_1$. Note that, with regularizers, the synthetic frame will still be visually sensible, or otherwise the 2D CNN pre-trained on ImageNet cannot be utilized for recognition. The performance of video recognition framework with different network architectures, or with different IFS variants, will be evaluated in the experimental section.

\section{Experiments}
\subsection{Datasets}
The experiments are conducted on Kinetics-400 \cite{carreira2017quo}, UCF101 \cite{UCF101} and HMDB51 \cite{HMDB51} datasets. IFS is optimized on the training set of Kinetics-400 and then is applied on all the dataset. \textbf{Kinetics-400} is one of the large-scale action recognition benchmarks. It consists of around 300K videos from 400 action categories. The 300K videos are divided into 240K, 20K, 40K for training, validation and test set, respectively. Each video in this dataset is 10-second short clip cropped from the raw YouTube video. Note that the labels for test set are not publicly available and the performances on Kinetics-400 dataset are all reported on the validation~set. After optimization, the effectiveness of synthetic frame is validated on Kinetics-400, UCF101 and HMDB51 in the context of video recognition. \textbf{UCF101} and \textbf{HMDB51} are two of the most popular video action recognition benchmarks. UCF101 consists of 13,320 videos from 101 action categories, and HMDB51 consists of 6,849 videos from 51 action categories. Each split in UCF101 includes about 9.5K training and 3.7K test videos, while a HMDB51 split contains 3.5K training and 1.5K test videos. We report the average results over three splits on these two datasets.

\subsection{Network Architecture}
The detailed structures of encoder-decoder networks ($\mathcal{F}$, $\mathcal{F}_{a}^{-1}$, $\mathcal{F}_{m}^{-1}$) and encoder networks ($\mathcal{C}$, $\mathcal{D}$) are given in Table~\ref{tab:arch}. The encoder-decoder networks originate from the 9-block ResNet in \cite{zhu2017unpaired}, which shows promising results on image-to-image translation task. This network contains two down-scale convolutional layers, two up-scale deconvolutional layers and nine residual blocks. For the encoder network, we devise the N-Layer classifier in \cite{zhu2017unpaired}, which stacks five $4\times 4$ convolutional layers. Therefore, the encoder network can produce a feature map with $7\times 7$ resolution which can be utilized to categorize the synthetic frame ($\mathcal{C}$) or distinguish between the real frame and synthetic frame ($\mathcal{D}$).

\begin{table}
\centering
\small
\caption{\small The detailed architectures of encoder and encoder-decoder networks in IFS.}
\vspace{0.1cm}
\begin{tabularx}{0.47\textwidth}{@{~~}X@{~~}|@{~~}c@{~~}|@{~~}c@{~~}|@{~~}c@{~~}} \toprule
\multirow{2}{*}{\textbf{Layer~~}} & \textbf{Encoder} & \textbf{Encoder-Decoder} & \multirow{2}{*}{\textbf{Output Size}}\\
& \textbf{Network} & \textbf{Network} & \\ \midrule
\multirow{2}{*}{conv1} & \multirow{2}{*}{4$\times$4, s=2} & \multirow{2}{*}{7$\times$7} & \emph{En} : $32\times112^{2}$  \\
& & & \emph{En-De} : $32\times224^{2}$  \\ \midrule
\multirow{2}{*}{conv2} & \multirow{2}{*}{4$\times$4, s=2} & \multirow{2}{*}{4$\times$4, s=2} & \emph{En} : $64\times56^{2}$  \\
& & & \emph{En-De} : $64\times112^{2}$  \\ \midrule
\multirow{2}{*}{conv3} & \multirow{2}{*}{4$\times$4, s=2} & \multirow{2}{*}{4$\times$4, s=2} & \emph{En} : $128\times28^{2}$  \\
& & & \emph{En-De} : $128\times56^{2}$  \\ \midrule
\multirow{2}{*}{conv4} & \multirow{2}{*}{4$\times$4, s=2} & \multirow{2}{*}{${\begin{bmatrix} \text{3}\times\text{3}\\ \text{3}\times\text{3} \end{bmatrix}}_{\text{res}}\times \text{9}$} & \emph{En} : $256\times14^{2}$  \\
& & & \emph{En-De} : $128\times56^{2}$  \\ \midrule
\multirow{2}{*}{conv5} & \multirow{2}{*}{4$\times$4, s=2} & 4$\times$4, s=1/2 & \emph{En} : $512\times7^{2}$  \\
& & 4$\times$4, s=1/2 & \emph{En-De} : $32\times224^{2}$  \\ \bottomrule
\end{tabularx}
\vspace{-0.05in}
\label{tab:arch}
\end{table}

\subsection{Training and Inference Strategy}
The proposed IFS is implemented on PyTorch framework with multiple GPUs in parallel. We use MPEG-4 encoded videos, which have on average 11 P-frames for every I-frame. For the \textbf{training of IFS} (Section \ref{sec:opt}), we set the clip size as $T\times 224 \times 224$, where $T=12$. The clips are randomly cropped without overlapping and is resized with the short edge in $[256, 340]$. IFS synthesizes one frame for average 12-frame clip. The clips are randomly flipped along horizontal direction for data augmentation. The network parameters are optimized by Adam \cite{kingma2014adam} method with ${\beta}_{1}=0.9$, ${\beta}_{2}=0.999$. The learning rate is initially set to 0.001, which is annealed down to zero following a cosine decay. The size of mini-batch is 128 and the optimization will be completed after 64 epoches. The optimized network $\mathcal{F}$ will be utilized for frame synthesis. For \textbf{video recognition} (Section \ref{sec:vr}), we utilize the same data pre-processing. The initial learning rate is 0.01 and the weights are optimized by standard stochastic gradient descent. Each mini-batch contains either 256 frames for 2D CNN or 256 clips for 3D CNN and the training completes after 256 epochs. In the inference stage, we employ the three-crop strategy in \cite{feichtenhofer2019slowfast} and average the scores from 20/10 uniformly sampled synthetic frames/clips for 2D/3D CNN, respectively. All the variants of IFS follow exactly the same training and inference strategy as IFS.

\begin{table}[t]
    \centering
    \small
    \caption{\small Top-1 classification accuracy on Kinetics-400 with the synthetic frames end-to-endly generated by IFS variants. The variants are trained with the task of appearance reconstruction plus different combinations of regularizers. The last column "jpeg" shows the performance when the synthetic frames are compressed as JPEG images.}
    \vspace{0.1cm}
    \begin{tabularx}{0.40\textwidth}{X|c|c|cc} \toprule
    	\multirow{2}{*}{\textbf{Task}} & \multirow{2}{*}{\textbf{Regularizer}} &  \multirow{2}{*}{$\mathcal{L}_{app}$} & \multicolumn{2}{c}{\textbf{Top-1}}\\
        & & & end-to-end & jpeg \\ \midrule
        \multirow{3}{*}{$app$} & -- & 0.002 & 71.7 & 59.2 \\
         & $adv$ & 0.014 & 71.2 & 70.3 \\
         & $adv+color$ & 0.014 & \textbf{72.8} & \textbf{72.8} \\ \bottomrule
    \end{tabularx}
    \label{tab:adv_color}
\end{table}

\begin{figure}[!tb]
   \centering {\includegraphics[width=0.47\textwidth]{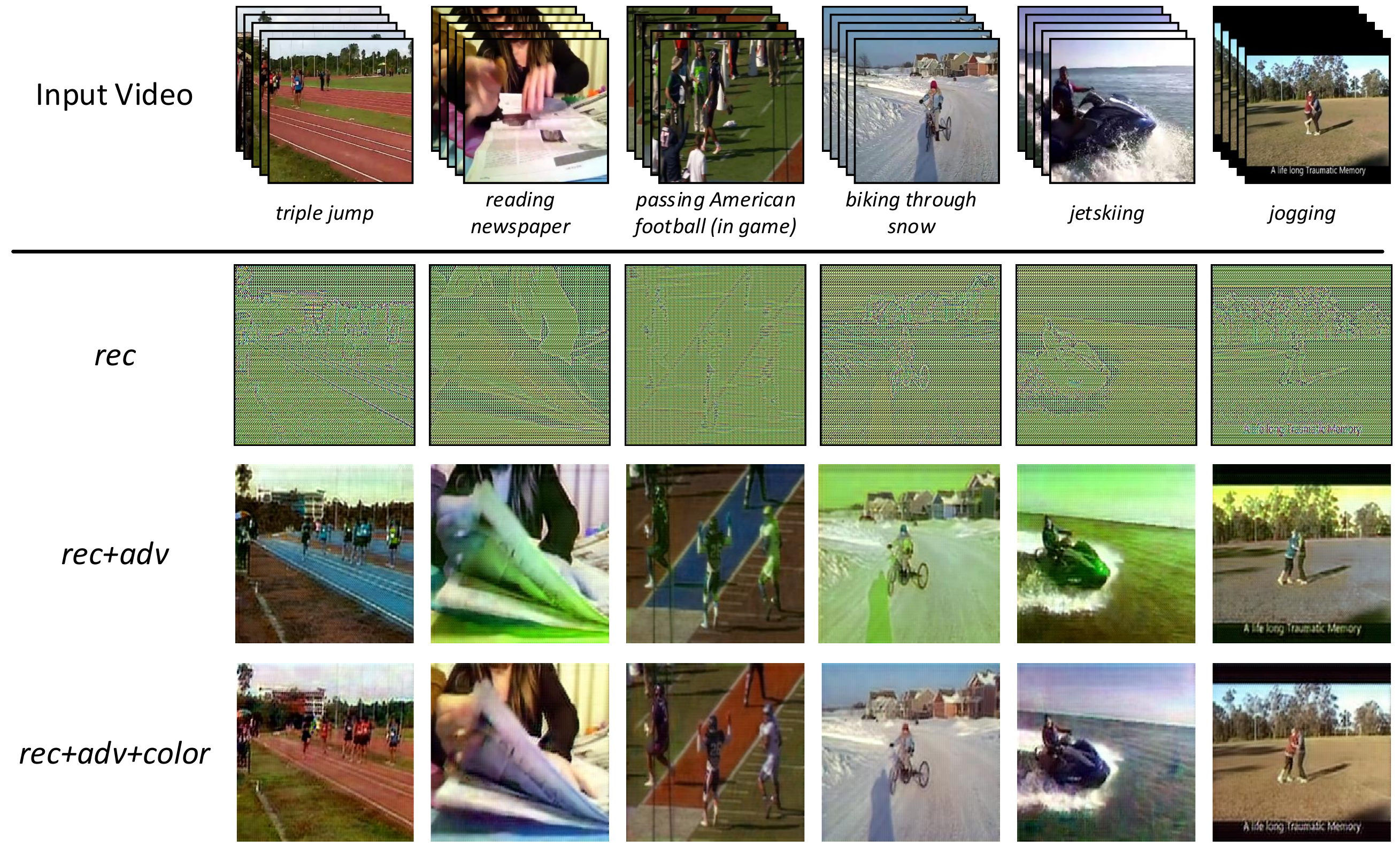}}
   \caption{\small The examples of synthetic frames generated by IFS trained with the reconstruction task and different regularizers.}
   \label{fig:example_quality}
   \vspace{-0.2in}
\end{figure}

\subsection{Evaluations on the Regularizers}
We first examine the impact of two regularizers, i.e., adversarial loss ($adv$) and color consistency ($color$) on IFS training. For simplicity, the IFS is only trained with the reconstruction task, i.e. Equ. (\ref{eq:loss-app}).
Table \ref{tab:adv_color} summarizes the top-1 classification accuracies along with the loss when different regularizers are trained together. Figure \ref{fig:example_quality} further contrasts the synthetic frames to show the impact of regularizers on visual effect. Without regularizer, the quality of synthetic frames is inexplicable despite the low value in reconstruction loss. The top-1 accuracy achieves 71.7\% when a synthetic frame, represented as a matrix of floating point values, is input to 2D CNN. As expected, the accuracy decreases to 59.2\% when these synthetic frames are compressed as JPEG images. By taking adversarial learning ($adv$) into account, the synthetic frames are more visually realistic. The accuracy does not drop dramatically when the frames are compressed as JPEG images. Nevertheless, this results in an increase of reconstruction loss and a drop of classification accuracy on the frame without compression. When color consistency is further considered, the accuracy reaches 72.8\% for both the original and compressed synthetic frames. While a synthetic frame is not necessarily required to be visually realistic, a frame with decent visual quality can fully leverage the pre-trained network learnt from images (ImageNet) for video recognition. Making the synthetic frame visually similar to the real frame can ensure the effective transfer learning of the pre-trained network, especially when the input images are compressed to save storage space. In the rest of the paper, we compress the synthetic frames as JPEG images to reduce the demand for disk space. The time complexity of IFS and disk space consumption of storing JPEG images will be discussed in the supplementary material.

\begin{table}[t]
    \centering
    \small
    \caption{\small The top-1 accuracy on Kinetics-400 of video classification with different combinations of tasks.}
\vspace{0.1cm}
    \begin{tabularx}{0.47\textwidth}{l|CCC|c} \toprule
    	\textbf{Task} & $\mathcal{L}_{app}$ & $\mathcal{L}_{cat}$ & $\mathcal{L}_{mot}$ & \textbf{Top-1}\\ \midrule
        $app$ & 0.014 & -- & -- & 72.8\\
        $cat$ & -- & 3.263 & & 68.2\\
        $mot$& -- & -- & 0.007  & 71.6\\ \midrule
        $app+cat$& 0.017 & 3.652 & --  & 73.7 \\
        $app+mot$& 0.015 & -- & 0.010  & 74.5 \\
        $cat+mot$& -- & 3.340 & 0.008 & 72.8 \\ \midrule
        $app+cat+mot$& 0.015 & 3.701 & 0.011 & \textbf{75.0} \\ \bottomrule
    \end{tabularx}
    \label{tab:rec_cat_mot}
   \vspace{-0.05in}
\end{table}

\begin{figure*}[!tb]
   \centering {\includegraphics[width=0.98\textwidth]{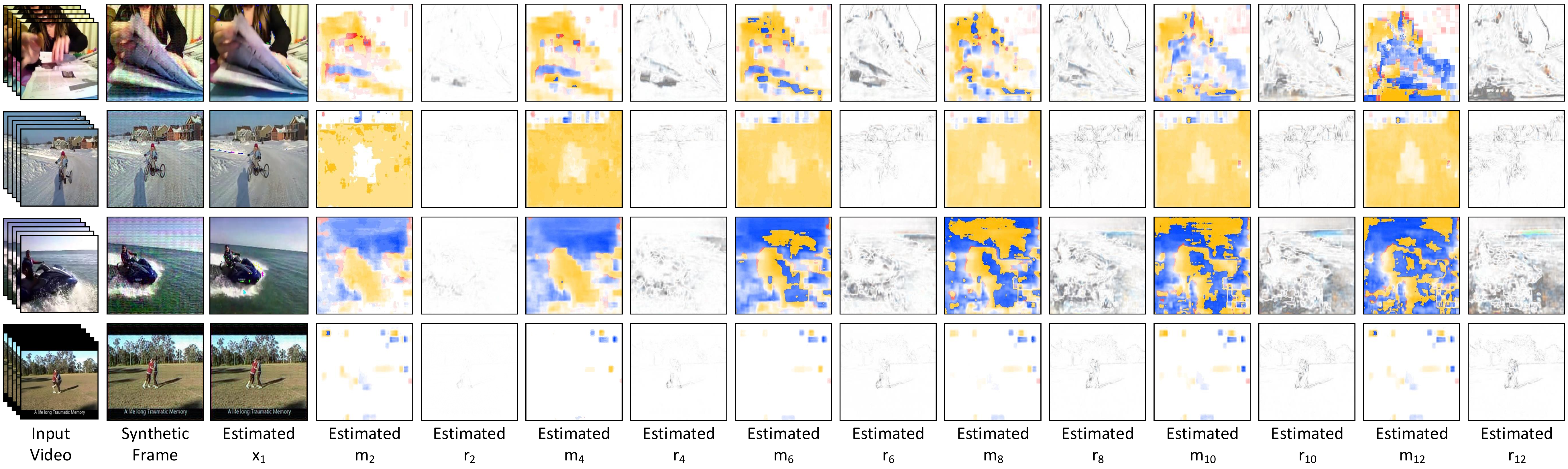}}
   \caption{\small Examples of the input video, the synthetic frame and the estimated I-frame, motion vectors and residuals.}
   \label{fig:example_decode}
\end{figure*}

\subsection{Evaluations on the Tasks}
\begin{figure}[!tb]
   \centering {\includegraphics[width=0.47\textwidth]{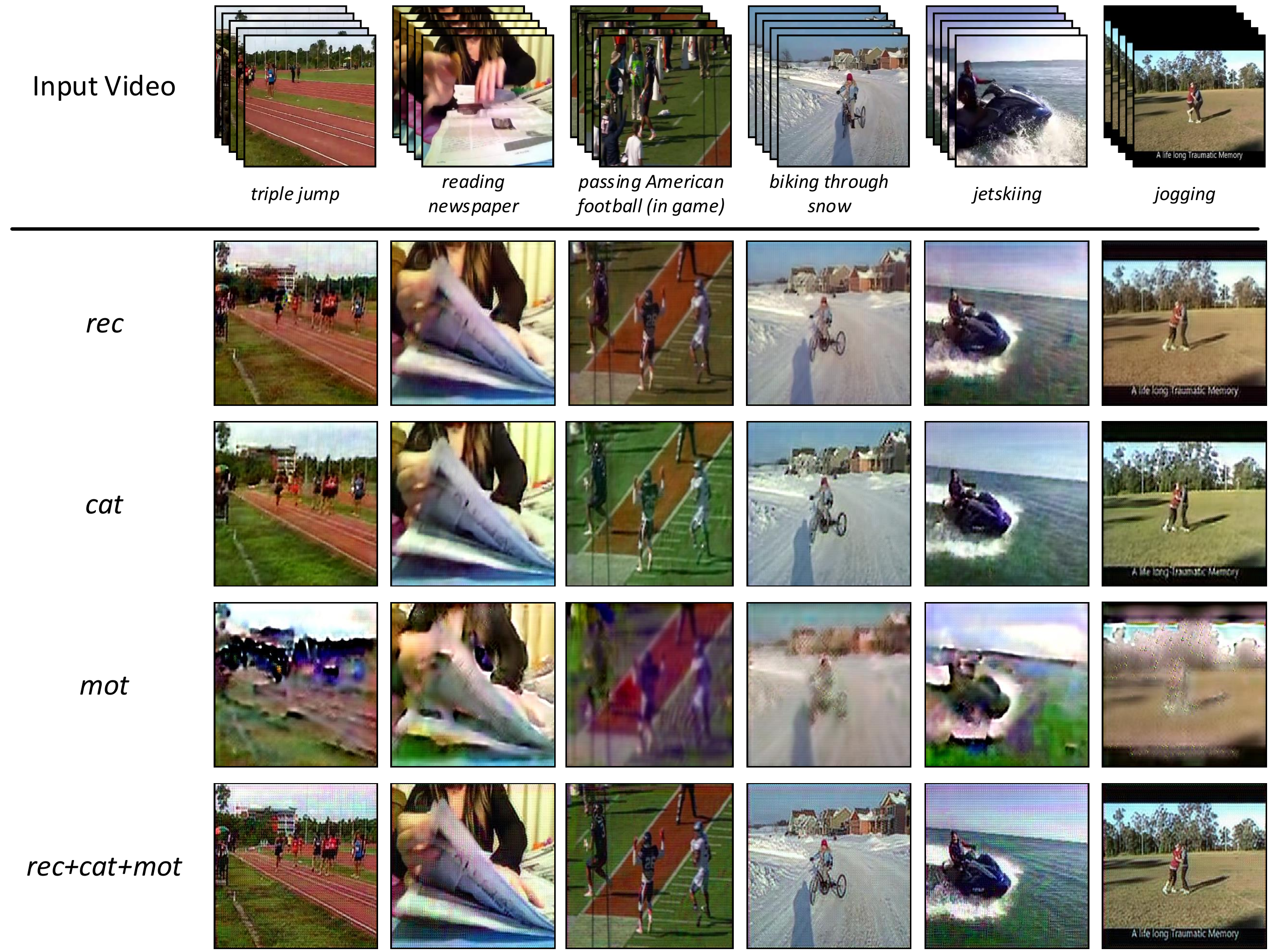}}
   \caption{\small Examples of the synthetic frames generated by different tasks from the input sequences.}
   \label{fig:example_content}
\vspace{-0.2in}
\end{figure}

Next, we study how video classification is affected by different tasks. Table \ref{tab:rec_cat_mot} details the losses of the optimization on different tasks and the top-1 video classification accuracy. Figure \ref{fig:example_content} shows the examples of the input video sequences and their synthetic frames via different tasks.
Note that both regularizers are trained together with the tasks. Among the three tasks, the reconstruction task exhibits the highest performance and the result is much better than the top-1 accuracy attained by the video categorization task. This somewhat reveals the weakness of frame distillation via categorization task, where the emphasis is on the image-level semantics rather than the pixel-level supervision as in the appearance reconstruction task and motion estimation task.
Further improvement in classification is noted when combining multiple tasks for joint training. The highest result is attained when all the three tasks are involved in training. Figure \ref{fig:example_decode} visualizes the frames synthesized by IFS for various videos, along with the reconstructed I-frames, estimated motion maps and residuals. The result verifies that the motion dynamics and visual details can be encapsulated simultaneously in a 2D frame by our IFS framework.

\subsection{Evaluations on Informative Frame Synthesis}
Next, we verify that the rich information (i.e., visual details and motion dynamics) captured in a 2D synthetic frame is helpful for video classification. We compare against the results when the raw data such as I-frames, motion vectors and residuals are directly extracted from MPEG videos for classification. Furthermore, other approaches such as optical flow, AVE, DI and SVMP are also compared as shown in Table \ref{tab:rec_IFS}. The \textbf{optical flow} modality firstly computes the value of optical flow between consecutive frames with \cite{zach2007duality}, and then converts to a 2-channel image by scaling the value to [0, 255]. \textbf{AVE} is the average ``frame" over all the frames. \textbf{DI} and \textbf{SVMP} applies approximate rank pooling \cite{bilen2016dynamic} and SVM pooling \cite{wang2018video}, respectively, on the raw pixels of the frames, and takes decision boundary as the informative frame. AVE, DI and SVMP are all utilized to transform a 12-frame clip to a single frame. \textbf{IFS} and \textbf{IFS-mot} executes our framework with all three tasks and without appearance reconstruction task, respectively.

\begin{table}[t]
    \centering
    \small
    \caption{\small Performance comparisons on Kinetics-400 of video classification with the informative frame obtained via different ways.}
\vspace{0.1cm}
    \begin{tabularx}{0.42\textwidth}{X|c|cc} \toprule
    	\multirow{2}{*}{\textbf{Method}} & \multirow{2}{*}{\textbf{One-stream}} & \multicolumn{2}{c}{\textbf{Two-stream}} \\
        & & \textbf{+I-frame} & \textbf{+flow} \\ \midrule
        I-frame & 72.7 & -- & 74.6 \\
        motion vector & 34.0 & 73.4 & 65.2 \\
        residual & 69.2 & 74.3 & 73.1 \\ \midrule
        optical flow & 63.2 & 74.6 & -- \\
        AVE & 70.2 & 73.7 & 73.8 \\
        DI \cite{bilen2016dynamic} & 71.6 & 74.9 & 72.9 \\
        SVMP \cite{wang2018video} & 71.0 & 74.4 & 72.3\\ \midrule
        \textbf{IFS} & \textbf{75.0} & \textbf{75.4} & \textbf{77.3} \\
        \textbf{IFS}-mot & 72.8 & 74.8 & 74.2 \\ \bottomrule
    \end{tabularx}
    \label{tab:rec_IFS}
   \vspace{-0.05in}
\end{table}

Table \ref{tab:rec_IFS} lists the top-1 accuracy of video classification on Kinetics-400. Overall, the results across one-stream and two-stream evaluations consistently indicate that IFS leads to a performance boost against other baselines. An interesting observation is that DI and SVMP are superior to AVE, but still inferior to I-frame. We speculate that this is due to the lack of local structure and details of the output frame by AVE, DI and SVMP. In contrast, by encapsulating visual details and temporal evolution in 2D frame synthesis, IFS exhibits better performance.
Further improvement can be attained if fusing the synthetic frame with either I-frame or optical flow. The result shows that the synthetic frame indeed complements well with other modalities. Note that when the appearance reconstruction task is not included in IFS (i.e., IFS-mot), the performance is also considerably better than the other motion-only modalities such as optical flow and motion vector.

\begin{table}[t]
    \centering
    \small
    \caption{\small Performance comparison with the state-of-the-art methods on Kinetics-400, in terms of accuracy and computational complexity measured in GFLOPs $\times$ views. The number of views represents the number of clips sampled from the full video during inference, i.e., the number of temporal samples $\times$ the number of spatial crops. ``N/A'' indicates the numbers are not available for~us.}
\vspace{0.1cm}
    \begin{tabularx}{0.47\textwidth}{X@{~}|@{~}c@{~}|@{~~}c@{~~}c} \toprule
    	\textbf{Method} & \textbf{GFLOPs$\times$views} & \textbf{top-1} & \textbf{top-5} \\ \midrule
        STM R50 \cite{jiang2019stm} & 66$\times$30 & 73.7 & 91.6 \\
        DFB R152 \cite{martinez2019action} & N/A$\times$200 & 74.3 & 91.4 \\
        AVD R101 \cite{tavakolian2019avd} & 10$\times$N/A & 69.9 & 92.5 \\ \midrule
        I3D \cite{carreira2017quo} & 108$\times$N/A & 72.1 & 90.3 \\
        R(2+1)D \cite{tran2018closer} & 152$\times$115 & 72.0 & 90.0 \\
        S3D-G \cite{xie2018rethinking} & 143$\times$N/A & 77.2 & 93.0 \\
        Non-local R101 \cite{wang2018non} & 359$\times$30 & 77.7 & 93.3 \\
        LGD-3D \cite{qiu2019learning} & 195$\times$30 & 79.4 & 94.4 \\
        DFB R152-3D \cite{martinez2019action} & N/A$\times$N/A & 78.8 & 93.6 \\
        irCSN \cite{tran2019video} & 96.7$\times$30 & 79.0 & 93.5 \\
        X3D-XL \cite{feichtenhofer2020x3d} & 48.4$\times$30 & 79.1 & 93.9 \\
        SlowFast 8$\times$8 \cite{feichtenhofer2019slowfast} & 106$\times$30 & 77.9 & 93.2 \\
        SlowFast 16$\times$8 \cite{feichtenhofer2019slowfast} & 213$\times$30 & 78.9 & 93.5 \\
        SlowFast 16$\times$8+NL \cite{feichtenhofer2019slowfast} & 234$\times$30 & 79.8 & 93.9 \\ \midrule
        Two-stream AVD R101 \cite{tavakolian2019avd} & 21$\times$N/A  & 75.1 & 93.4 \\
        Two-stream I3D \cite{carreira2017quo} & 216$\times$N/A & 75.7 & 92.0 \\
        Two-stream R(2+1)D \cite{tran2018closer} & 304$\times$115 & 73.9 & 90.9 \\
        Two-stream LGD-3D \cite{qiu2019learning} & 390$\times$30 & 81.2 & 95.2 \\ \midrule

        I-frame & 10$\times$60 & 72.7 & 89.5 \\
        I-frame+flow & 21$\times$60 & 74.6 & 91.3\\
        \textbf{IFS} & 10$\times$60 & 75.0 & 91.5 \\
        \textbf{IFS}+\textbf{IFS}-mot & 21$\times$60 & 77.2 & 92.9 \\
        \textbf{IFS}-3D & 98$\times$30 & 79.0 & 94.0 \\
        \textbf{IFS}-3D+\textbf{IFS}-mot-3D & 196$\times$30 & 80.5 & 94.9 \\ \bottomrule
    \end{tabularx}
    \label{tab:sota}
    \vspace{-0.2in}
\end{table}

\subsection{Comparison with the State-of-the-art}
We compare with several state-of-the-art architectures in the context of video classification on Kinetics-400 dataset and Table \ref{tab:sota} summarizes the performance comparison. When using 2D CNN for video classification, IFS and IFS+IFS-mot show better performance than I-frame and I-frame+flow, respectively, which directly exploit the intra-coded frame or optical flow in the sequence. Our IFS also outperforms AVD R101 \cite{tavakolian2019avd} which synthesizes the frame by solely capitalizing on adversarial learning. Such result basically indicates the advantage of exploring multi-task learning plus two regularizers in IFS. Furthermore, IFS with less GFLOPs is even superior to several 3D CNN, e.g., I3D \cite{carreira2017quo} and R(2+1)D \cite{tran2018closer}, which spend about ten times GFLOPs. When taking 3D CNN as the architecture for video recognition, IFS-3D with less computation exhibits better performance than S3D-G \cite{xie2018rethinking} and Non-local \cite{wang2018non}. Despite sharing similar computational load, IFS-3D outperforms SlowFast 8$\times$8 \cite{feichtenhofer2019slowfast}. The performance of IFS-3D is below LGD-3D \cite{qiu2019learning} but the computation of IFS-3D is only half of LGD-3D. The two-stream integration of IFS-3D+IFS-mot-3D reaches the top-1 accuracy of 80.5\%, which is also higher than that of AVD R101, I3D, and R(2+1)D in two-stream mode.

\begin{table}[t]
    \centering
    \small
    \caption{\small Comparison with state-of-the-art on UCF101\&HMDB51.}
\vspace{0.1cm}
    \begin{tabularx}{0.47\textwidth}{X|c@{~}c|c@{~~}c} \hline
    	\textbf{Method} & \textbf{+Flow} & \textbf{+Kinetics} & \textbf{U101} & \textbf{H51} \\ \toprule
        IDT \cite{wang2013action} & & & 86.4 & 61.7 \\
        Two-stream \cite{simonyan2014two} & \checkmark & & 88.0 & 59.4 \\
        TSN \cite{wang2016temporal} & \checkmark & & 94.2 & 69.4 \\ \midrule
        I3D \cite{carreira2017quo} & & \checkmark & 95.4 & 74.5 \\
        S3D \cite{xie2018rethinking} & & \checkmark & 96.8 & 75.9 \\
        LGD-3D \cite{qiu2019learning} & & \checkmark & 97.0 & 75.7 \\
        STM \cite{jiang2019stm} & & \checkmark & 96.2 & 72.2 \\ \midrule
        AVD \cite{tavakolian2019avd} & \checkmark & \checkmark & 97.3 & 77.1 \\
        I3D \cite{carreira2017quo} & \checkmark & \checkmark & 97.9 & 80.2 \\
        R(2+1)D \cite{tran2018closer} & \checkmark & \checkmark & 97.3 & 75.9 \\
        LGD-3D \cite{qiu2019learning} & \checkmark & \checkmark & 98.2 & 80.5 \\ \midrule
        \textbf{IFS}-3D & & \checkmark & 97.4 & 76.2 \\
        \textbf{IFS}-3D+\textbf{IFS}-mot-3D & & \checkmark & 98.2 & 80.3 \\ \bottomrule
    \end{tabularx}
    \label{tab:ucf_hmdb}
    \vspace{-0.2in}
\end{table}

We finally evaluate the transferability of video representation learnt by IFS plus video classification networks on UCF101 and HMDB51 datasets. Specifically, we first pre-train both IFS network and 3D CNN of IFS-3D and IFS-mot-3D on Kinetics-400 dataset, and then fine-tune on UCF101 and HMDB51. Following \cite{wang2016temporal}, we freeze the parameters of all Batch Normalization layers except for the first one and add an extra dropout layer with 0.9 dropout rate to reduce the effect of over-fitting. Table \ref{tab:ucf_hmdb} shows the performance comparisons. With only RGB input, IFS-3D pre-trained on Kinetics leads to better performance than I3D and STM \cite{jiang2019stm}. When fusing two IFS variants, IFS-3D+IFS-mot-3D, without optical flow extraction, performs better than the two-stream R(2+1)D, AVD and I3D.

\section{Conclusions}
This paper explores knowledge distillation from video sequence to frame for activity recognition. Particularly, we study the problem from a novel viewpoint of early fusing a 3D video clip to a 2D informative frame. To materialize our idea, we have devised Informative Frame Synthesis (IFS) architecture which integrates three objective tasks and two regularizers. Each task and regularizer empowers a synthetic frame to capture specific information ranging from visual to motion for video classification. Extensive experiments conducted on Kinetics dataset validate each design in IFS. The results of video classification by IFS with 2D CNN or 3D CNN on three video datasets demonstrate a good compromise between classification and computation cost. Furthermore, the ability in abstracting the knowledge from video as just-one-frame is potentially a new paradigm of video processing. IFS has demonstrated the feasibility of using such synthesized frames for video understanding.

\textbf{Acknowledgments.} This work was supported by the National Key R\&D Program of China under Grant No. 2020AAA0108600.

{\small
\bibliographystyle{ieee_fullname}
\bibliography{egbib}
}

\end{document}